\documentclass{article}

\usepackage{PRIMEarxiv}

\usepackage[utf8]{inputenc} 
\usepackage[T1]{fontenc}    
\usepackage{hyperref}       
\usepackage{url}            
\usepackage{booktabs}       
\usepackage{amsfonts}       
\usepackage{nicefrac}       
\usepackage{microtype}      
\usepackage{lipsum}
\usepackage{fancyhdr}       
\usepackage{graphicx}       
\graphicspath{{media/}}     
\title{AutoViVQA: A Large-Scale Automatically Constructed Dataset
  for Vietnamese Visual Question Answering
}

\author{
  Nguyen Anh Tuong$^{1,2}$,
  Phan Ba Duc$^{1,2}$,
  Nguyen Trung Quoc$^{1,2}$,\\
  Tran Dac Thinh$^{1,2}$,
  Dang Duy Lan$^{1,2}$,
  Nguyen Quoc Thinh$^{1,2}$,
  Tung Le$^{1,2}$\\[6pt]
  $^{1}$Faculty of Information Technology,
  University of Science, VNU-HCM, Vietnam\\[4pt]
  $^{2}$Vietnam National University, Ho Chi Minh City, Vietnam\\[4pt]
  \begin{minipage}{\textwidth}
    \centering
    {\small\texttt{\{22120071, 22120182, 22120301,
                    22120347, 22120348, 22120412\}}}\\
    {\small\texttt{@student.hcmus.edu.vn}}\\[2pt]
    {\small\texttt{lttung@fit.hcmus.edu.vn}}
  \end{minipage}
}

\begin{document}
\maketitle

\begin{abstract}
Visual Question Answering (VQA) is a fundamental multimodal task that
requires models to jointly understand visual and textual information.
Early VQA systems relied heavily on language biases, motivating
subsequent work to emphasize visual grounding and balanced datasets.
With the success of large-scale pre-trained transformers for both text
and vision domains---such as PhoBERT for Vietnamese language
understanding and Vision Transformers (ViT) for image representation
learning---multimodal fusion has achieved remarkable progress.

For Vietnamese VQA, several datasets have been introduced to promote
research in low-resource multimodal learning, including ViVQA,
OpenViVQA, and the recently proposed ViTextVQA. These resources enable
benchmarking of models that integrate linguistic and visual features in
the Vietnamese context.

Evaluation of VQA systems often employs automatic metrics originally
designed for image captioning or machine translation, such as BLEU,
METEOR, CIDEr, Recall, Precision, and F1-score. However, recent
research suggests that large language models can further improve the
alignment between automatic evaluation and human judgment in VQA tasks.

In this work, we explore Vietnamese Visual Question Answering using
transformer-based architectures, leveraging both textual and visual
pre-training while systematically comparing automatic evaluation
metrics under multilingual settings.
\end{abstract}

\section{Introduction}



Visual Question Answering (VQA) is a multimodal task in which a model must interpret an image, understand a natural-language question, and generate an accurate answer. As a benchmark, VQA evaluates a system’s ability to integrate visual perception with linguistic and commonsense reasoning, making it a central problem in multimodal artificial intelligence. Recent advances in English-centric VQA have been driven by large-scale multimodal models such as BLIP-2~\cite{li2023blip2}, Flamingo~\cite{alayrac2022flamingo}, and LLaVA~\cite{liu2023visualinstructiontuning}, which demonstrate strong capabilities in visual grounding, reasoning, and open-ended response generation. However, these advances remain largely inaccessible to low-resource languages.

Vietnamese, despite being spoken by nearly 100 million people and increasingly used in AI-driven applications, still lacks large-scale and high-quality multimodal benchmarks for VQA. Existing Vietnamese VQA datasets, including ViVQA~\cite{tran2021vivqa} and OpenViVQA~\cite{nguyen2023openvivqa}, suffer from several limitations. Their scale is often insufficient to support training or adaptation of modern multimodal models, while question diversity remains limited, with an emphasis on object-centric or text-centric queries. More complex reasoning phenomena such as multi-step inference, spatial relations, causal reasoning, and culturally grounded interpretation are underrepresented. In addition, data quality is inconsistent: manual annotation is expensive and difficult to scale, whereas naive AI-assisted generation frequently introduces hallucinations, weak visual grounding, and cultural or social biases.

These limitations reveal a broader methodological challenge beyond dataset size: how to construct Vietnamese VQA data that is scalable, reasoning-aware, and quality-controlled without heavy reliance on human annotation. To address this challenge, we propose \textbf{AutoViVQA}, a large-scale Vietnamese VQA dataset constructed through a \textbf{LLM-driven generation and quality assurance framework}. Rather than focusing solely on data collection, our approach formulates dataset construction as a controlled generation process. It explicitly regulates reasoning complexity through a multi-level criteria schema and applies an ensemble-based validation protocol to filter noisy or weakly grounded samples.

As a result, AutoViVQA achieves substantially greater scale and diversity than existing Vietnamese VQA resources, while maintaining strong visual grounding and linguistic naturalness. Beyond the dataset itself, the proposed framework provides a reproducible methodology for building high-quality VQA benchmarks in low-resource languages and enables systematic evaluation of multimodal models under controlled reasoning conditions.



\paragraph{Contributions.}
This work makes the following contributions:

\begin{itemize}
    \item We introduce \textbf{AutoViVQA}, Vietnamese Visual Question Answering dataset constructed entirely through a \textbf{LLM-driven pipeline}, addressing the scarcity of high-quality multimodal benchmarks for low-resource languages.
    
    \item We propose a \textbf{quality assurance VQA generation framework} that explicitly regulates question types and cognitive complexity through a five-level reasoning schema, enabling balanced coverage of recognition, relational, compositional, causal, and text-in-image reasoning.
    
    \item We design an \textbf{ensemble-based validation protocol} that combines multi-model evaluation, criterion-wise thresholding, and majority voting to automatically filter noisy or weakly grounded samples without relying on human annotation.
\end{itemize}




\section{Related Work}\label{sec:related-work}

Approaches on synthetic data generation have expanded rapidly with the advancement of large language models. Existing studies investigate instruction induction, multimodal data augmentation, and human-aligned criteria for improving dataset quality. These efforts demonstrate the potential of synthetic data for scaling supervised tasks, yet they are largely concentrated on English and have limited exploration in low-resource languages, including Vietnamese.

Instruction generation is a key component of synthetic data research. Prior works create new task formulations from small seed sets, enabling large language models to generalize across diverse task types \cite{zhou2023largelanguagemodelshumanlevel,zhang2023autoinstructautomaticinstructiongeneration}. Our approach builds on this literature by comparing instructions from multiple models and community-curated prompts, followed by the use of genre-specific constraints to shape question structure and reasoning style. These candidate instructions are evaluated on two thousand samples, filtered, and selected through a voting mechanism to ensure linguistic naturalness and diversity. Synthetic data generation and augmentation form another major research direction \cite{honovich2022unnaturalinstructionstuninglanguage,guo2024generativeaisyntheticdata,meng2023tuninglanguagemodelstraining,yang-etal-2020-generative}. 

Although prior work demonstrates that large language models can enhance coverage and robustness, applications to multimodal tasks such as Visual Question Answering remain limited. Our study extends this line of research by focusing specifically on VQA, incorporating focus-aware prompting and OCR-guided signals to better align visual content with the generated questions and answers. Recognizing that synthetic multimodal data may still lack semantic depth or reliable grounding, we incorporate targeted augmentation and refinement steps to enhance conceptual richness and linguistic naturalness. This combination produces VQA data with stronger multimodal consistency and improved reasoning complexity. Our work contributes a novel direction for multimodal dataset construction by unifying structured instruction design, systematic prompt evaluation, and vision-aware generation into an automated pipeline tailored for Vietnamese VQA. This integrated framework addresses key limitations in existing synthetic data research and enables the creation of a large-scale, culturally grounded, and automatically verified Vietnamese VQA dataset.

\section{AutoViVQA Data Pipeline}

\subsection{Dataset Overview}

We present AutoViVQA, a Vietnamese Visual Question Answering dataset constructed through a systematic, multi-stage pipeline designed to maximize both annotation quality and reasoning diversity. The dataset integrates real-world visual contexts from MS COCO \cite{lin2015coco} with high-quality Vietnamese textual captions from the Vista\-vi\_llava\_complex\_reasoning corpus \cite{ViVLM_Vista_2024}, ensuring natural language fidelity and strong image-text alignment.


\begin{figure}[ht!]
    \centering
    \includegraphics[width=0.6\linewidth]{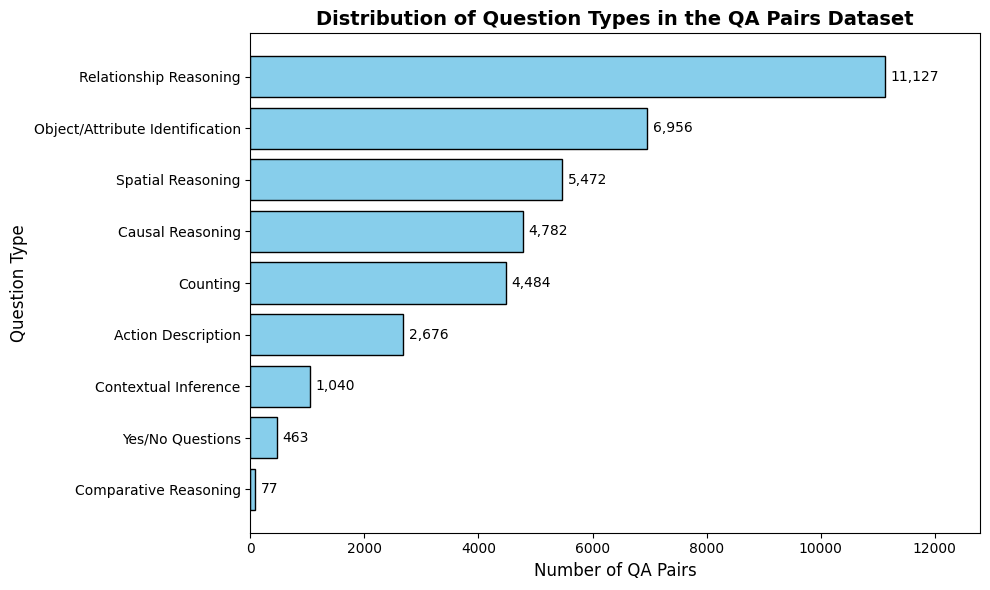}
    \caption{Distribution of Question Types in the QA Pairs Dataset}
    \label{fig:qa_pairs_hist}
\end{figure}

\noindent AutoViVQA contains 19,411 images, 37,077 questions, and 185,385 answers, with five answers per question. The dataset is divided into an eighty percent training split and a twenty percent validation split. Questions are short, typically between one and ten words, and encourage natural, human-like responses. As illustrated in Figure~\ref{fig:qa_pairs_hist}, AutoViVQA covers nine question categories and five levels of reasoning complexity, resulting in a collection that captures recognition, relational understanding, multi-step reasoning, causal inference, and text-in-image interpretation. This structure is particularly important for Vietnamese, a language that exhibits rich tonal patterns, flexible compound formation, and context-dependent semantic ambiguity.

\begin{figure}[ht!]
    \centering
    \includegraphics[width=1\linewidth]{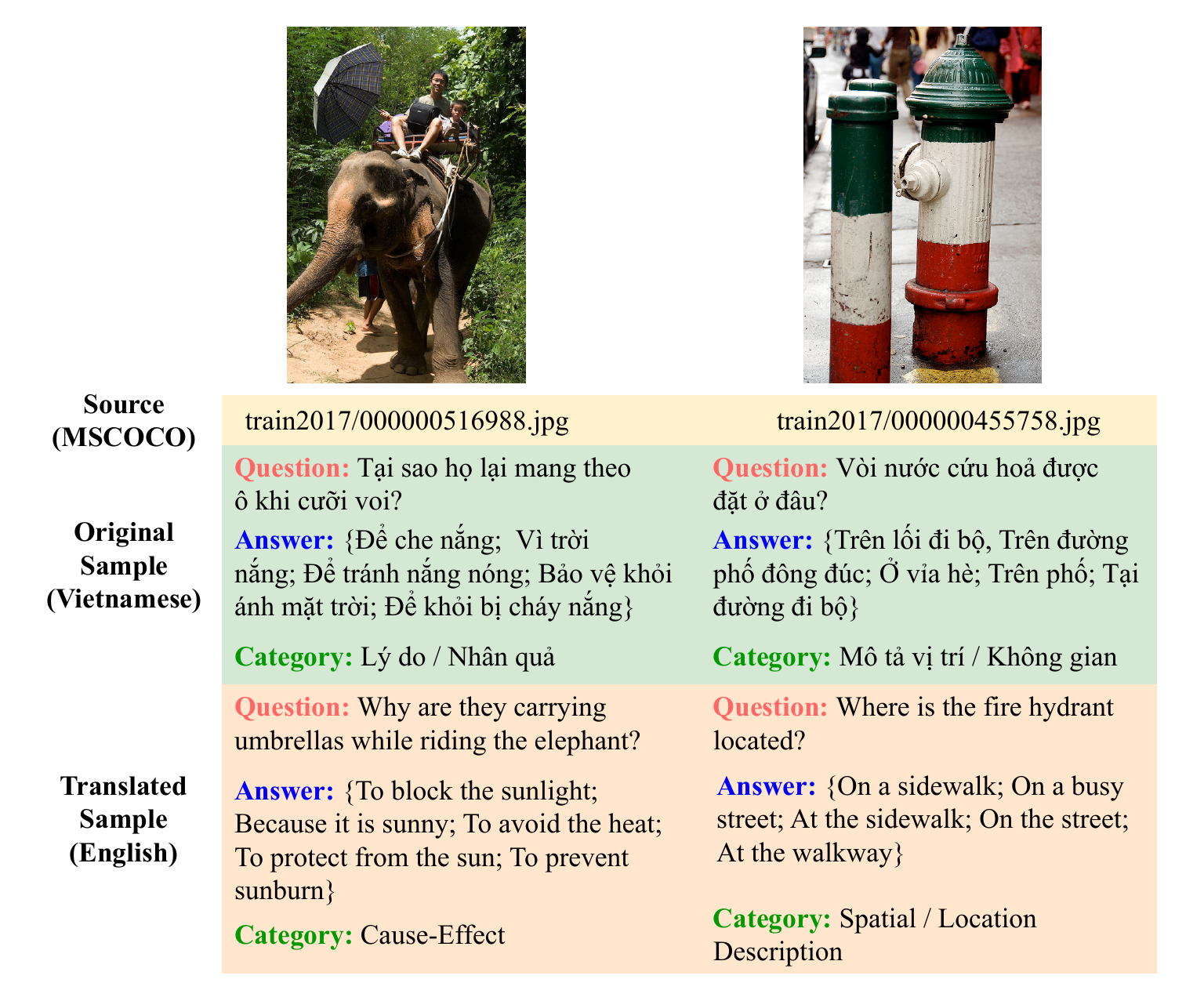}
    \caption{Example image–question–answer pairs from the AutoViVQA dataset, illustrating the diversity of scenes, question types, and reasoning levels.}
    \label{fig:dataset_examples}
\end{figure}

Figure~\ref{fig:dataset_examples} presents examples of image, question, and answer pairs. These samples demonstrate the diversity of visual scenes and the range of reasoning skills required by the dataset. Each question is accompanied by five independent answers, enabling models to be evaluated using consensus-based scoring similar to established VQA benchmarks.

\begin{table*}[htbp]
\centering
\renewcommand{\arraystretch}{1.25}
\setlength{\tabcolsep}{4pt}
\caption{Comparison of AutoViVQA with existing English and Vietnamese VQA datasets.}
\label{tab:dataset_comparison}
\resizebox{\textwidth}{!}{
\begin{tabular}{p{2.5cm}|p{2.8cm}|p{2.8cm}|p{2.8cm}|p{2.8cm}|p{2.8cm}|p{2.8cm}}
\hline
\textbf{Criteria} &
\textbf{AutoViVQA (Ours)} &
\textbf{VQA v2 \cite{goyal2017makingvqa}} &
\textbf{OpenViVQA \cite{nguyen2023openvivqa}} &
\textbf{ViCLEVR \cite{tran2023viclevr}} &
\textbf{ViOCRVQA \cite{pham2024viocrvqa}} &
\textbf{ViTextVQA \cite{nguyen2025vitextvqa}} \\ 
\hline
\textbf{Language} &
Vietnamese (Vi) &
English (En) &
Vietnamese (Vi) &
Vietnamese (Vi) &
Vietnamese (Vi) &
Vietnamese (Vi) \\ 
\hline
\textbf{Images} &
19,411 (MS COCO) &
204,721 (MS COCO) &
11,199 (Vietnamese images) &
26,216 (Synthetic images) &
28,282 (Book covers) &
16,762 (Scene text) \\ 
\hline
\textbf{QA Pairs} &
37,077 Q / 185,385 A (5 A/Q) &
1.1M Q / 11M A &
37,914 Q / 37,914 A &
30,000 Q / 30,000 A &
123,781 Q / 123,781 A &
50,342 Q / 50,342 A \\ 
\hline
\textbf{Annotation Method} &
Semi-automatic (LLM / Gemini-2.5-Flash) &
Manual annotation &
Manual annotation (Crowd-sourcing) &
Semi-automatic (Grammar-based) &
Semi-automatic (Metadata / Template-based) &
Manual annotation \\ 
\hline
\textbf{Question Types} &
9 focused types (Relation, Causal, Spatial, Counting, etc.) &
Mainly 3 basic types (Yes/No, Number, Other) &
Open (Text QA \& Non-text QA) &
6 diagnostic types (Color, Size, Shape, Material, Count, Comparison) &
5 focused types (Title, Author, Publisher, Translator, Genre) &
Text-based QA (Scene text only) \\ 
\hline
\textbf{Reasoning Level} &
5 levels (Compositional, Commonsense/Causal, Text-in-Image) &
General reasoning &
Open / Semantic reasoning &
Visual reasoning (Compositional) &
Text understanding (OCR-VQA) &
Scene text comprehension \\ 
\hline
\textbf{Answer} &
Short (1–10 words, natural) &
Very short (89.41\% single words) &
Longer (avg. 6.9 words; 36\% full sentences) &
Short (mostly single words) &
Medium (avg. 7.52 words; >90\% phrases) &
Medium/Long (depends on OCR text) \\ 
\hline
\textbf{Balance} &
Rebalanced (Low Yes/No; High Relation/Causal) &
Balanced (reduced language bias) &
Reduced bias via open-ended format &
Diagnostic (even distribution across 6 Q types) &
Specialized (focused on 5 book-related aspects) &
Specialized (focused only on scene text) \\ 
\hline
\end{tabular}}
\end{table*}

Table~\ref{tab:dataset_comparison} presents a detailed comparison between AutoViVQA and existing English and Vietnamese VQA datasets. Beyond the dataset itself, AutoViVQA introduces a flexible LLM-driven construction framework. By adjusting prompts or reasoning templates, researchers can customize datasets for specialized domains, vary reasoning depth, or modify annotation styles. This extensibility highlights the role of AutoViVQA not only as a benchmark but also as a reproducible methodology for creating future Vietnamese multimodal datasets.

\textbf{Comparison with Foundational Benchmarks:} English VQA datasets such as VQA v2 \cite{goyal2017makingvqa} contain significantly larger quantities of data, but their question formats remain coarse and nearly ninety percent of their answers are single tokens \cite{nguyen2023openvivqa}. AutoViVQA introduces a more detailed annotation design that includes nine question types and five reasoning levels, supporting the development of models that can engage with relational, causal, and contextual understanding rather than simple classification.

\textbf{Comparison with Vietnamese Benchmarks}: Vietnamese VQA resources remain limited and often fragmented. Early Vietnamese datasets such as ViVQA relied on machine translation, which affects linguistic precision \cite{nguyen2025vitextvqa}. Subsequent datasets such as OpenViVQA \cite{nguyen2023openvivqa} expanded scale but contained long, less constrained answers. Other datasets focus on narrow domains, including the synthetic reasoning scenes in ViCLEVR \cite{tran2023viclevr}, scene-text oriented questions in ViTextVQA \cite{nguyen2025vitextvqa}, and book-cover based queries in ViOCRVQA \cite{pham2024viocrvqa}. AutoViVQA advances this landscape by providing a real-world Vietnamese VQA benchmark that combines natural images, high-quality Vietnamese text, and a structured reasoning framework. The dataset is produced using a semi-automatic process with Gemini 2.5 Flash, making it both scalable and adaptable.

To maintain evaluation consistency, AutoViVQA adopts concise answers composed of between one and ten tokens. This design preserves the precision of classification-style VQA and avoids the overly long responses common in other open-ended Vietnamese datasets, such as OpenViVQA~\cite{nguyen2023openvivqa}.


\subsection{Data Collection}
Given the design objectives described above, the construction of AutoViVQA begins with assembling reliable multi-modal resources that provide both visual diversity and linguistically faithful Vietnamese descriptions. This stage establishes the foundation upon which the subsequent data creation and pre-processing components operate.

\begin{figure}[ht!]
\centering
\includegraphics[width=\linewidth]{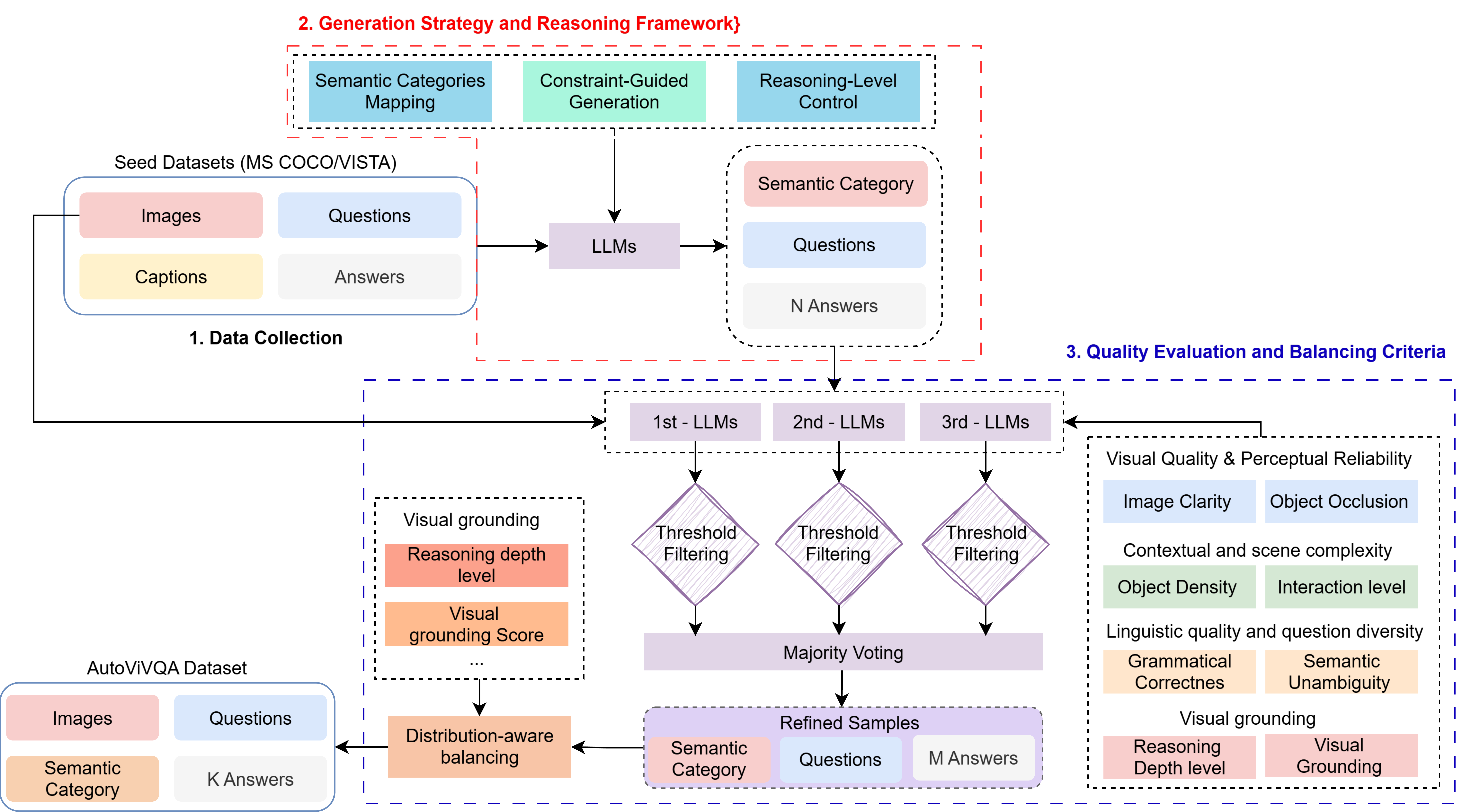}
\caption{AutoViVQA generation and quality control framework combining semantic-guided LLM generation, multi-LLM evaluation, and distribution-aware dataset balancing}
\label{fig:dataset_pipeline}
\end{figure}

The dataset is constructed from carefully curated visual and textual resources to ensure both semantic alignment and linguistic accuracy. We integrate real-world images from MS COCO \cite{lin2015coco} with Vietnamese captions and conversational descriptions from VISTA \cite{ViVLM_Vista_2024}. This combination provides the visual diversity and linguistic richness necessary for generating reasoning-focused Vietnamese VQA samples. After identifying overlapping samples across sources, we apply the pipeline illustrated in Figure~\ref{fig:dataset_pipeline}, where candidate English samples are translated, expanded, and filtered through our Quality Controller to produce the final high-quality AutoViVQA dataset.

Each image is associated with several caption variants and conversation examples. These multilayer descriptions enrich the semantic space by capturing different aspects of the same scene. Table~\ref{tab:VISTA-examples} illustrates how captions and conversations complement one another, such as in kayak scenes that emphasize proximity, activity, and perspective, or elephant scenes that highlight actions and environmental context.


\begin{table}[ht!]
\caption{Representative examples from the VISTA dataset showing conversation and caption diversity}\label{tab:VISTA-examples}
\centering
\small
\begin{tabular}{p{0.25\textwidth}p{0.35\textwidth}p{0.35\textwidth}}
\toprule
\textbf{Scene} & \textbf{Caption Variants} & \textbf{Conversation Examples} \\
\midrule
Kayak on water & 
Two kayaks on calm water; Kayakers enjoying peaceful lake; Watercraft near shoreline &
Q: Which kayak is closer to shore? A: The red one on the left; Q: What activity are people doing? A: Kayaking \\
\midrule
Elephant and tree & 
Elephant reaching for leaves; Wildlife feeding behavior; Tree interaction scene &
Q: What is the elephant doing? A: Eating leaves from tree; Q: Is this in the wild? A: Yes \\
\bottomrule
\end{tabular}
\end{table}

The integration of COCO and VISTA establishes a well-grounded multimodal corpus for Vietnamese VQA generation. The visual diversity of COCO and the linguistic quality of VISTA jointly support the creation of questions that reflect natural Vietnamese usage and span a broad range of reasoning skills. This multimodal foundation is critical for the generation of high-quality Vietnamese VQA annotations in later stages of the pipeline.



\subsection{Generation Strategy and Reasoning Framework}
This stage introduces a reasoning-aware framework for automatically generating structured Vietnamese VQA annotations from multimodal inputs. The proposed approach is not intended to introduce a new VQA model architecture, but to formalize a controllable and verifiable data generation process that decomposes dataset construction into explicit reasoning supervision, semantic constraints, and distributional control. Large language models are employed as constrained generators rather than autonomous decision-makers.

The overall design is guided by two principles: (i) explicit control over reasoning complexity during generation, and (ii) post-hoc verifiability through structured outputs and distributional monitoring. This formulation enables systematic analysis and reproducibility, addressing limitations of unconstrained LLM-based data synthesis.

\subsubsection{Constraint-Guided Generation}
To generate high-quality question--answer pairs, we employ a controlled prompting strategy that integrates semantic grounding and reasoning-level constraints. For each image, all available caption variants are merged into a unified textual context, providing diverse but bounded semantic cues that guide the model toward generating contextually faithful Vietnamese questions.

The model is explicitly instructed with a target reasoning level and produces one question accompanied by five short answers (approximately ten words each). This multi-answer design simulates responses from multiple annotators and supports downstream consensus-based validation, facilitating the identification of ambiguous or difficult cases during quality control. The prompting constraints enforce three requirements: (i) outputs must follow natural Vietnamese syntax, (ii) questions must remain strictly grounded in the provided captions without introducing unsupported details, and (iii) answers must be logically consistent with the question intent, such as color, number, spatial relation, or causal explanation.

Importantly, the prompting mechanism serves as an interface for enforcing reasoning and semantic constraints, rather than constituting the core contribution itself.


\subsubsection{Reasoning-Level Control}
We define a five-level reasoning schema to explicitly regulate the cognitive depth of generated questions. The levels span recognition, spatial and relational understanding, compositional reasoning, commonsense and causal inference, and text-in-image interpretation:
\begin{itemize}
    \item Level 1 (Recognition): Identifying objects or basic attributes
    \item Level 2 (Spatial \& Relational): Reasoning about spatial relationships or simple comparisons
    \item Level 3 (Compositional): Multi-step reasoning involving multiple objects or actions
    \item Level 4 (Commonsense \& Causal): Inferring intentions, mental states, or causal relationships
    \item Level 5 (Text-in-Image): Reading and interpreting textual content within the image
\end{itemize}

The dataset is balanced to approximate a normal distribution across reasoning levels, with target proportions of 0.05, 0.24, 0.40, 0.24, and 0.05 for Levels~1 to~5, respectively. During generation, deviations from these targets are continuously monitored, and levels exhibiting the largest deficits are preferentially selected. This mechanism can be interpreted as a curriculum-style control over reasoning depth, preventing collapse toward trivial recognition-based questions.


\subsubsection{Semantic Category Mapping}
Each generated question is assigned to a single semantic category that reflects its dominant reasoning operation, including object or attribute identification, spatial description, action understanding, numerical reasoning, comparisons, relational inference, causal analysis, and contextual inference. These categories are directly associated with the predefined reasoning levels, as summarized in Table~\ref{tab:categories}.

\begin{table}[h]
\centering
\caption{Mapping of question categories to reasoning depth levels.}
\label{tab:categories}
\begin{tabular}{ll}
\toprule
\textbf{Category} & \textbf{Reasoning Levels} \\
\midrule
Object / Attribute Identification & Level 1 \\
Location / Spatial Description    & Level 2 \\
Action Description                & Levels 2–3 \\
Counting                          & Levels 1–3 \\
Yes/No Question                   & Levels 1–5 \\
Comparisons                       & Levels 2–5 \\
Relationships                     & Levels 2–5 \\
Causal Reasoning                  & Levels 4–5 \\
Contextual Inference              & Levels 4–5 \\
\bottomrule
\end{tabular}
\end{table}

When multiple categories correspond to a selected reasoning level, those requiring deeper reasoning, such as relational or causal questions, are prioritized. If the caption lacks sufficient semantic cues to support the chosen level, the framework selects the next feasible level with a positive deficit. This constrained category assignment restricts the semantic hypothesis space and ensures that all generated questions remain well-grounded and cognitively interpretable.

Overall, this framework can be viewed as a reasoning-controlled data generation process that integrates explicit supervision over cognitive depth, semantic scope, and distributional balance. By decomposing generation into interpretable control variables, the proposed approach moves beyond ad-hoc prompting and provides a reproducible foundation for constructing large-scale Vietnamese VQA datasets.

\subsection{Quality Evaluation and Balancing Criteria}
To ensure dataset quality, diversity, and reasoning coverage, we define a structured set of automatic evaluation criteria spanning four dimensions: visual quality, contextual complexity, linguistic validity, and multimodal reasoning demand. All criteria are estimated automatically using large language models and vision--language models, without human annotation.

Visual quality and perceptual reliability measure whether images provide sufficient evidence for question answering, using image clarity, object occlusion, and discriminability between similar entities. These criteria serve as hard filters to remove visually ambiguous samples. Contextual and scene complexity is quantified through object density, interaction level, and scene clutter, and is later used for balancing the dataset across different difficulty levels.

Linguistic quality and question diversity are evaluated based on grammatical correctness, semantic unambiguity, QA structural validity, syntactic diversity, question-type distribution, and language naturalness. A bias sensitivity score is additionally employed to suppress subjective or culturally biased questions. Samples failing minimum linguistic thresholds are discarded, while diversity-related criteria guide balancing.

To explicitly enforce multimodal grounding, we introduce a Visual Grounding Score that measures whether a question genuinely requires visual information to answer, following the principle of making the visual modality essential. Furthermore, each sample is annotated with a reasoning depth level spanning recognition, spatial/relational, compositional, commonsense/causal, and text-in-image reasoning. All criteria are normalized to a unified scale and aggregated using data-driven threshold-based filtering and distribution-aware balancing.

\subsection{Post-generation Quality Control}
After generating the initial Vietnamese VQA annotations, we apply a fully automatic post-generation quality control pipeline that operationalizes the evaluation criteria described above to ensure visual reliability, linguistic validity, and balanced reasoning coverage.

All generated samples are first normalized into a unified tabular schema and subjected to basic sanity checks to remove invalid, duplicated, or malformed question--answer pairs. This step discarded 167 out of 60,000 samples ($\sim$0.28\%), which is within acceptable bounds for large-scale generative pipelines. Key semantic attributes, such as scene types and primary object references, are additionally canonicalized to reduce lexical fragmentation. Statistical summaries over automatically estimated quality scores are then computed, and median-based aggregation is used to derive data-driven reference thresholds.

Reliability filtering is performed using an ensemble-based evaluation framework. Each sample is assessed by an ensemble of $2n+1$ independent vision--language and language models, each producing scores over eighteen quality dimensions grouped into four categories: visual quality, contextual complexity, linguistic quality, and visual grounding (including the Visual Grounding Score) with reasoning depth. For each dimension, the dataset median is used as a non-parametric threshold to binarize scores, and final decisions are obtained via majority voting across the ensemble. A sample is retained only if it satisfies at least nine of the eighteen criteria, enforcing holistic quality while preserving diversity. This filtering stage reduces the dataset to approximately 37,000 high-quality instances.

Finally, distribution-aware balancing is applied to prevent dominance of frequent or trivial categories. Samples are weighted toward visual grounding strength and reasoning depth, while labels with extremely low support are merged or removed. Controlled under-sampling ensures that category frequencies differ by no more than ten percent, yielding a dataset that is both quality-controlled and suitable for robust Vietnamese VQA benchmarking.

\section{Experiments}

This section evaluates the effectiveness of the proposed data construction and ensemble-based filtering framework, with a particular focus on its impact on downstream Vietnamese Visual Question Answering and vision-conditioned text generation tasks. Rather than introducing new model architectures, our experiments are designed to assess whether higher-quality, reasoning-aware data produced by our pipeline leads to more stable, accurate, and semantically grounded model behavior under controlled settings.

\subsection{Experimental Setup}

All experiments are conducted on a Vietnamese vision--language benchmark derived from the AutoViVQA dataset, where each sample consists of an image, a Vietnamese question, and a corresponding answer. The objective of these experiments is to evaluate the effectiveness of the proposed data construction and ensemble-based filtering framework, rather than to introduce new model architectures.

The dataset is produced through a multi-stage refinement pipeline driven by large language models, including language identification, grammatical assessment, semantic consistency verification, factuality validation, and length normalization. The refined corpus is split into training, validation, and test sets using an $8{:}1{:}1$ ratio, and all models are trained and evaluated on the same filtered splits to ensure fair comparison.

We benchmark a diverse set of models spanning Vietnamese-specific vision--language systems (Vintern (base)~\cite{doan2024vintern1b}, Vintern (finetuned), ViT5\_ViT, BARTPhoBEiT~\cite{tran2023bartphobeitpretrainedsequencetosequenceimage}), general-purpose large language models (GPT-5, LLaMA~3.2), and commercial multimodal models (Gemini~2.0~Flash, Gemini~2.5~Flash). These models serve as probes to analyze how different architectures respond to the same high-quality training distribution.

To isolate the impact of dataset quality, all models are trained under matched optimization settings and evaluated using default inference configurations, without model-specific prompt tuning or decoding adjustment. Performance is assessed using standard automatic metrics, including Accuracy, Precision, Recall, F1, BLEU, ROUGE-L, METEOR, and CIDEr, capturing complementary aspects of generation quality and grounding.

\subsection{Main Results}

Table~\ref{tab:main_results} summarizes the performance of all evaluated models when trained and tested on the same refined dataset produced by our framework. Since no model-specific prompt tuning or decoding optimization is applied, the reported results primarily reflect the impact of data quality and filtering strategies rather than architectural or hyperparameter differences.

\begin{table}[h!]
\centering
\small
\caption{Performance comparison on the Vietnamese text generation test set.
Best results per column are in bold.}
\label{tab:main_results}
\begin{tabular}{lcccccccc}
\toprule
Model & Acc & Prec & Rec & F1 & BLEU & ROUGE & METEOR & CIDEr \\
\midrule
Vintern (base) & 0.0012 & 0.1752 & 0.1987 & 0.1755 & 0.0191 & 0.2584 & 0.2393 & 0.0854 \\
\textbf{Vintern (finetune)} & \textbf{0.1301} & \textbf{0.5247} & 0.5512 & \textbf{0.5376} & 0.0611 & \textbf{0.5193} & \textbf{0.3525} & 0.7284 \\
ViT5\_ViT & 0.0797 & 0.4684 & 0.5033 & 0.4852 & 0.0413 & 0.4689 & 0.3102 & 0.7268 \\
BARTPhoBEiT & 0.0881 & 0.4530 & 0.4648 & 0.4588 & \textbf{0.4329} & 0.4483 & 0.2457 & \textbf{1.8896} \\
GPT 5 & 0.1084 & 0.4720 & 0.5520 & 0.5089 & 0.0607 & 0.4730 & 0.3334 & 0.8420 \\
LLaMA 3.2 & 0.0036 & 0.2396 & 0.7371 & 0.3616 & 0.0362 & 0.3611 & 0.3001 & 0.6284 \\
Gemini 2.0 F & 0.0055 & 0.2720 & 0.7410 & 0.3979 & 0.0441 & 0.3960 & 0.3172 & 0.7442 \\
Gemini 2.5 F & 0.0022 & 0.2443 & \textbf{0.7666} & 0.2475 & 0.0039 & 0.3727 & 0.3122 & 0.7190 \\
\bottomrule
\end{tabular}
\end{table}

\noindent Across model families, we observe consistent improvements in precision-oriented and semantic fidelity metrics, including Precision, Recall, F1, ROUGE-L, METEOR, and CIDEr. These trends indicate that the proposed refinement pipeline produces cleaner and more semantically grounded supervision signals, leading to more stable and controlled generation behavior in downstream Vietnamese vision--language tasks.

To isolate the effect of dataset refinement under a fixed architecture, we compare Vintern (base) with its finetuned counterpart trained on the filtered corpus. Under identical architectural settings, training on the refined dataset yields an approximately threefold improvement in F1 and an eightfold increase in CIDEr. These substantial gains demonstrate that the proposed filtering framework effectively removes noisy or weakly grounded samples and enhances semantic alignment, rather than introducing improvements through model-specific modifications.

Beyond this controlled case study, similar patterns emerge across other model families. Vietnamese-specific vision--language models generally benefit most from the refined dataset, exhibiting higher precision and stronger semantic consistency. In contrast, general-purpose and commercial large language models tend to achieve higher Recall but comparatively lower Precision, suggesting a trade-off between output diversity and semantic control. This behavior is consistent with their open-ended generation characteristics and highlights the importance of reasoning-aware data filtering in reducing hallucination and improving grounding.

Overall, the results confirm that the proposed data construction and ensemble-based validation framework yields quantitatively significant and consistent improvements across heterogeneous architectures. Importantly, these gains arise from systematic enhancements in dataset quality rather than from model design choices, underscoring the generality and reusability of the proposed approach for building high-quality Vietnamese vision--language benchmarks.



\subsection{Human Validation and Error Analysis}

To complement the automatic ensemble-based validation, we conduct a targeted human validation study to assess the quality of the generated dataset. A subset of 1000 question--answer pairs is randomly sampled from AutoViVQA and evaluated by three bilingual Vietnamese annotators with experience in multimodal NLP tasks.

Each sample is assessed along three criteria: (i) linguistic fluency, (ii) semantic correctness with respect to the image, and (iii) appropriateness of the assigned reasoning level. Inter-annotator agreement is measured using Krippendorff’s alpha, yielding a score of $\alpha = 0.72$, which indicates substantial agreement and suggests that the evaluation criteria are well-defined and consistently interpreted.

Overall, the majority of evaluated samples are judged to be fluent, visually grounded, and aligned with their intended reasoning categories. These findings are consistent with trends observed in automatic validation, providing empirical support that the proposed ensemble filtering protocol effectively selects high-quality and semantically reliable data.

\paragraph{Error Analysis.}
Manual inspection of the annotated samples reveals four dominant residual error categories: (1) visually under-specified questions that admit multiple plausible interpretations, (2) overly generic answers lacking sufficient specificity, (3) mild cultural awkwardness in phrasing, and (4) rare hallucinated attributes not supported by the image. These cases account for fewer than 6\% of the inspected samples and are largely filtered out during the ensemble validation stage.

The remaining errors highlight inherent challenges in fully automated VQA dataset construction and motivate future improvements in reasoning-level calibration, cultural adaptation, and bias-aware filtering mechanisms.

\section{Conclusion}

This work presents \textbf{AutoViVQA}, a large-scale Vietnamese Visual Question Answering benchmark constructed through a fully automated, LLM-driven generation and validation framework. By combining reasoning-level controlled question generation with an ensemble-based quality validation protocol, AutoViVQA advances beyond existing Vietnamese VQA resources in terms of scale, question diversity, reasoning coverage, and visual grounding. Empirical evaluations demonstrate that datasets produced by our pipeline consistently improve downstream Vietnamese VQA and text generation performance across both language-specific and general-purpose multimodal models, highlighting the effectiveness of the proposed framework for low-resource settings.

Despite these advances, AutoViVQA has several limitations. The dataset is built on MS COCO images, which may restrict visual diversity in culturally specific Vietnamese contexts. Although the ensemble-based validation protocol substantially reduces noise and weak grounding, residual biases inherited from large language models may persist. In addition, AutoViVQA primarily reflects standard Vietnamese and does not explicitly model regional or dialectal variation. Addressing these limitations by incorporating more culturally diverse visual sources, stronger bias-aware validation mechanisms, and dialect-sensitive generation strategies constitutes an important direction for future work.

Overall, AutoViVQA provides not only a high-quality Vietnamese VQA benchmark but also a scalable and reproducible methodology for constructing reasoning-aware multimodal datasets in low-resource languages. We hope this work will facilitate more rigorous evaluation of multimodal models under controlled reasoning conditions and stimulate further research on culturally grounded and linguistically inclusive multimodal AI.

\section*{Acknowledgments}
This research is funded by the University of Science, VNU-HCM
under grant number CNTT 2025-02.

\bibliographystyle{unsrt}  
\bibliography{references}

@inproceedings{tran2021vivqa,
  title     = {ViVQA: Vietnamese Visual Question Answering},
  author    = {Tran, Khanh Quoc and Nguyen, An Trong and Le, An Tran-Hoai and Nguyen, Kiet Van},
  booktitle = {Proceedings of the 35th Pacific Asia Conference on Language, Information and Computation},
  year      = {2021},
  pages     = {683--691},
  url       = {https://aclanthology.org/2021.paclic-1.72/}
}

@article{nguyen2023openvivqa,
  title   = {OpenViVQA: Task, Dataset, and Multimodal Fusion Models for Visual Question Answering in Vietnamese},
  author  = {Nguyen, Nghia Hieu and Vo, Duong T. D. and Nguyen, Kiet Van and Nguyen, Ngan Luu-Thuy},
  journal = {Information Fusion},
  volume  = {100},
  pages   = {101868},
  year    = {2023},
  doi     = {10.1016/j.inffus.2023.101868}
}

@misc{nguyen2025vitextvqa,
  title  = {ViTextVQA: A Large-Scale Visual Question Answering Dataset for Evaluating Vietnamese Text Comprehension in Images},
  author = {Nguyen, Quan Van and Tran, Dan Quang and Pham, Huy Quang and Nguyen, Thang Kien-Bao and Nguyen, Nghia Hieu and Nguyen, Kiet Van and Nguyen, Ngan Luu-Thuy},
  year   = {2025},
  doi    = {10.48550/arXiv.2404.10652}
}

@inproceedings{goyal2017makingvqa,
  title     = {Making the V in VQA Matter: Elevating the Role of Image Understanding in Visual Question Answering},
  author    = {Goyal, Yash and Khot, Tejas and Summers-Stay, Douglas and Batra, Dhruv and Parikh, Devi},
  booktitle = {IEEE Conference on Computer Vision and Pattern Recognition (CVPR)},
  year      = {2017},
  pages     = {6325--6334},
  doi       = {10.1109/CVPR.2017.670}
}

@inproceedings{alayrac2022flamingo,
author = {Alayrac, Jean-Baptiste and Donahue, Jeff and Luc, Pauline and Miech, Antoine and Barr, Iain and Hasson, Yana and Lenc, Karel and Mensch, Arthur and Millicah, Katie and Reynolds, Malcolm and Ring, Roman and Rutherford, Eliza and Cabi, Serkan and Han, Tengda and Gong, Zhitao and Samangooei, Sina and Monteiro, Marianne and Menick, Jacob and Borgeaud, Sebastian and Brock, Andrew and Nematzadeh, Aida and Sharifzadeh, Sahand and Binkowski, Mikolaj and Barreira, Ricardo and Vinyals, Oriol and Zisserman, Andrew and Simonyan, Karen},
title = {Flamingo: a visual language model for few-shot learning},
year = {2022},
isbn = {9781713871088},
publisher = {Curran Associates Inc.},
address = {Red Hook, NY, USA},
booktitle = {Proceedings of the 36th International Conference on Neural Information Processing Systems},
articleno = {1723},
numpages = {21},
location = {New Orleans, LA, USA},
series = {NIPS '22}
}

@inproceedings{li2023blip2,
    title = {BLIP-2: bootstrapping language-image pre-training with frozen image encoders and large language models},
    author = {Li, Junnan and Li, Dongxu and Savarese, Silvio and Hoi, Steven},
    year = {2023},
    publisher = {JMLR.org},
    articleno = {814},
    numpages = {13},
    location = {Honolulu, Hawaii, USA},
    series = {ICML'23}
}

@inproceedings{liu2023visualinstructiontuning,
  title     = {Visual Instruction Tuning},
  author    = {Liu, Haotian and Li, Chunyuan and Wu, Qingyang and Lee, Yong Jae},
  booktitle = {Advances in Neural Information Processing Systems (NeurIPS)},
  year      = {2023},
  pages     = {34892--34916}
}

@misc{doan2024vintern1b,
  title  = {Vintern-1B: An Efficient Multimodal Large Language Model for Vietnamese},
  author = {Doan, Khang T. and Huynh, Bao G. and Hoang, Dung T. and Pham, Thuc D. and Pham, Nhat H. and Nguyen, Quan T. M. and Vo, Bang Q. and Hoang, Suong N.},
  year   = {2024},
  doi    = {10.48550/arXiv.2408.12480}
}

@misc{tran2023viclevr,
  title  = {ViCLEVR: A Visual Reasoning Dataset and Hybrid Multimodal Fusion Model for Visual Question Answering in Vietnamese},
  author = {Tran, Khiem Vinh and Phan, Hao Phu and Nguyen, Kiet Van and Nguyen, Ngan Luu-Thuy},
  year   = {2023},
  doi    = {10.48550/arXiv.2310.18046}
}

@misc{pham2024viocrvqa,
  title  = {ViOCRVQA: Novel Benchmark Dataset and Vision Reader for Visual Question Answering by Understanding Vietnamese Text in Images},
  author = {Pham, Huy Quang and Nguyen, Thang Kien-Bao and Nguyen, Quan Van and Tran, Dan Quang and Nguyen, Nghia Hieu and Nguyen, Kiet Van and Nguyen, Ngan Luu-Thuy},
  year   = {2024},
  doi    = {10.48550/arXiv.2404.18397}
}

@misc{tran2023bartphobeitpretrainedsequencetosequenceimage,
  title  = {BARTPhoBEiT: Pre-trained Sequence-to-Sequence and Image Transformers Models for Vietnamese Visual Question Answering},
  author = {Tran, Khiem Vinh and Nguyen, Kiet Van and Nguyen, Ngan Luu-Thuy},
  year   = {2023},
  doi    = {10.48550/arXiv.2307.15335}
}

@inproceedings{lin2015coco,
  title     = {Microsoft COCO: Common Objects in Context},
  author    = {Lin, Tsung-Yi and Maire, Michael and Belongie, Serge and Bourdev, Lubomir and Girshick, Ross and Hays, James and Perona, Pietro and Ramanan, Deva and Zitnick, C. Lawrence and Dollár, Piotr},
  booktitle = {European Conference on Computer Vision (ECCV)},
  year      = {2014},
  pages     = {740--755},
  doi       = {10.1007/978-3-319-10602-1_48}
}

@misc{zhou2023largelanguagemodelshumanlevel,
      title={Large Language Models Are Human-Level Prompt Engineers}, 
      author={Yongchao Zhou and Andrei Ioan Muresanu and Ziwen Han and Keiran Paster and Silviu Pitis and Harris Chan and Jimmy Ba},
      year={2023},
      eprint={2211.01910},
      archivePrefix={arXiv},
      primaryClass={cs.LG},
      url={https://arxiv.org/abs/2211.01910}, 
}

@article{ViVLM_Vista_2024,
  title={Vista},
  author={Tran, Oanh Ngoc and Bui, Hop Van and Ha, Hoang Huy and Phan, Phuc Van},
  year=2024,
  month={May},
  url={https://huggingface.co/datasets/Vi-VLM/Vista},
}

@misc{zhang2023autoinstructautomaticinstructiongeneration,
  title  = {Auto-Instruct: Automatic Instruction Generation and Ranking for Black-Box Language Models},
  author = {Zhang, Zhihan and Wang, Shuohang and Yu, Wenhao and Xu, Yichong and Iter, Dan and Zeng, Qingkai and Liu, Yang and Zhu, Chenguang and Jiang, Meng},
  year   = {2023},
  doi    = {10.48550/arXiv.2310.13127}
}

@misc{honovich2022unnaturalinstructionstuninglanguage,
  title  = {Unnatural Instructions: Tuning Language Models with (Almost) No Human Labor},
  author = {Honovich, Or and Scialom, Thomas and Levy, Omer and Schick, Timo},
  year   = {2022},
  doi    = {10.48550/arXiv.2212.09689}
}

@misc{guo2024generativeaisyntheticdata,
  title  = {Generative AI for Synthetic Data Generation: Methods, Challenges and the Future},
  author = {Guo, Xu and Chen, Yiqiang},
  year   = {2024},
  doi    = {10.48550/arXiv.2403.04190}
}

@misc{meng2023tuninglanguagemodelstraining,
  title  = {Tuning Language Models as Training Data Generators for Augmentation-Enhanced Few-Shot Learning},
  author = {Meng, Yu and Michalski, Martin and Huang, Jiaxin and Zhang, Yu and Abdelzaher, Tarek and Han, Jiawei},
  year   = {2023},
  doi    = {10.48550/arXiv.2211.03044}
}

@inproceedings{yang-etal-2020-generative,
  title     = {Generative Data Augmentation for Commonsense Reasoning},
  author    = {Yang, Yiben and Malaviya, Chaitanya and Fernandez, Jared and Swayamdipta, Swabha and Le Bras, Ronan and Wang, Ji-Ping and Bhagavatula, Chandra and Choi, Yejin and Downey, Doug},
  booktitle = {Findings of the Association for Computational Linguistics: EMNLP 2020},
  year      = {2020},
  pages     = {1008--1025},
  doi       = {10.18653/v1/2020.findings-emnlp.90}
}

\end{document}